\documentclass[11pt,a4paper]{article}

\usepackage[hyperref]{acl2020}
\usepackage{times}
\usepackage{latexsym}
\usepackage{arydshln}

\usepackage{amssymb}
\usepackage{pifont}
\usepackage[gen]{eurosym}
\usepackage{multirow}

\newcommand{\cmark}{\ding{51}}%
\newcommand{\xmark}{\ding{55}}%

\usepackage{microtype}

\aclfinalcopy 
\def\aclpaperid{437} 

\usepackage{comment}

\usepackage{amssymb}
\usepackage{amsmath}
\usepackage{mathtools}

\title{Translationese as a Language in ``Multilingual" NMT}

\author{Parker Riley$^\bigtriangleup$$^{*}$, Isaac Caswell$^\bigtriangledown$, Markus Freitag$^\bigtriangledown$, David Grangier$^\bigtriangledown$\\
$^\bigtriangleup$University of Rochester \\
$^\bigtriangledown$Google Research}
\date{}

\begin{document}
\maketitle

{\let\thefootnote\relax\footnotetext{*Work done while at Google Research.}}

\begin{abstract}
Machine translation has an undesirable propensity to produce ``translationese" artifacts, which can lead to higher BLEU scores while being liked less by human raters. Motivated by this, we model translationese and original (i.e.\ natural) text as separate languages in a multilingual model, and pose the question: can we perform zero-shot translation between original source text and original target text? There is no data with original source and original target, so we train a sentence-level classifier to distinguish translationese from original target text, and use this classifier to tag the training data for an NMT model. Using this technique we bias the model to produce more natural outputs at test time, yielding gains in human evaluation scores on both adequacy and fluency. Additionally, we demonstrate that it is possible to bias the model to produce translationese and game the BLEU score, increasing it while decreasing human-rated quality. We analyze these outputs using metrics measuring the degree of translationese, and present an analysis of the volatility of heuristic-based train-data tagging.
\end{abstract}

\section{Introduction}

``Translationese'' is a term that refers to artifacts present in text that was translated into a given language that distinguish it from text originally written in that language \citep{Gellerstam86}. These artifacts include lexical and word order choices that are influenced by the source language \citep{Gellerstam96} as well as the use of more explicit and simpler constructions \citep{Baker93}. 

These differences between translated and original text mean that the direction in which parallel data (bitext) was translated is potentially important for machine translation (MT) systems. Most parallel data is either source-original (the source was translated into the target) or target-original (the target was translated into the source), though sometimes neither side is original because both were translated from a third language.

\begin{figure}
    \centering
    \includegraphics[width=0.75\linewidth]{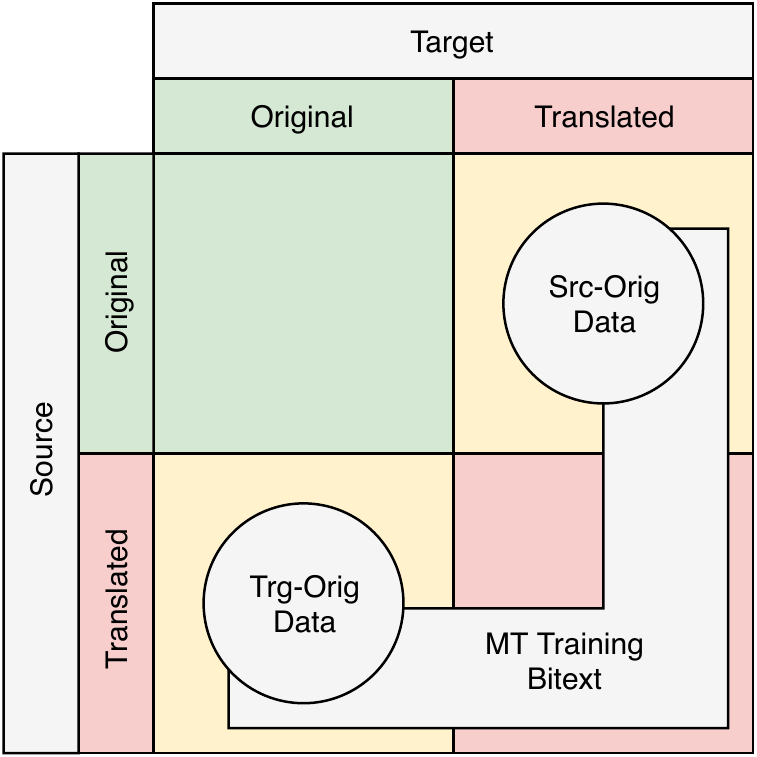}
    \caption{Illustration of MT train+test parallel data, organized into quadrants based on whether the source or target is translated or original.}
    \label{fig:grid_with_train}
    \vspace{-3mm}
\end{figure}
Figure~\ref{fig:grid_with_train} illustrates the four possible combinations of translated and original source and target data. Recent work has examined the impact of translationese in MT evaluation, using the WMT evaluation campaign as the most prominent example. From 2014 through 2018, WMT test sets were constructed such that 50\% of the sentence pairs are source-original (upper right quadrant of Figure~\ref{fig:grid_with_train}) and the rest are target-original (lower left quadrant).
\citet{Toral18}, \citet{Zhang19}, and \citet{Graham19} have examined the effect of this testing setup on MT evaluation, and have all argued that target-original test data should not be included in future evaluation campaigns because the translationese source is too easy to translate. While target-original test data does have the downside of a translationese source side, recent work has also shown that human raters prefer MT output that is closer in distribution to original target text than translationese \citep{Freitag19}. This indicates that the target side of test data should also be original (upper left quadrant of Figure~\ref{fig:grid_with_train}); however, it is unclear how to produce high-quality test data (let alone training data) that is simultaneously source- and target-original. 

Because of this lack of original-to-original sentence pairs, we frame this as a zero-shot translation task, where translationese and original text are distinct languages or domains. We adapt techniques from zero-shot translation with multilingual models \citep{johnson2016google}, where the training pairs are tagged with a reserved token corresponding to the domain of the target side: translationese or original text. Tagging is helpful when the training set mixes data of different types by allowing the model to 1) see each pair's type in training to preserve distinct behaviors and avoid regressing to a mean/dominant prediction across data types, and 2) elicit different behavior in inference, i.e.\ providing a tag at test time yields predictions resembling a specific data type.
We then investigate what happens when the input is an original sentence in the source language and the model's output is also biased to be original, a scenario never observed in training.

Tagging in this fashion is not trivial, as most MT training sets do not annotate which pairs are source-original and which are target-original\footnote{Europarl \citep{Koehn05} is a notable exception, but it is somewhat small and not in the news domain.}, so in order to distinguish them we train binary classifiers to distinguish original and translated target text.

Finally, we perform several analyses of tagging these ``languages" and demonstrate that tagged back-translation \citep{Caswell19} can be framed as a simplified version of our method, and thereby improved by targeted decoding.

Our contributions are as follows:
\begin{enumerate}
    \item We propose two methods to train translationese classifiers using only monolingual text, coupled with synthetic text produced by machine translation.
    \item Using only original$\to$translationese and translationese$\to$original training pairs, we apply techniques from zero-shot multilingual MT to enable original$\to$original translation.
    \item We demonstrate with human evaluations that this technique improves translation quality, both in terms of fluency and adequacy.
    \item We show that biasing the model to instead produce translationese outputs inflates BLEU scores while harming quality as measured by human evaluations.
    
\end{enumerate}
\section{Classifier Training + Tagging}
 
Motivated by prior work detailing the importance of distinguishing translationese from original text \citep{Kurokawa09,Lembersky12adapting,Toral18,Zhang19,Graham19,Freitag19,Edunov19} as well as work in zero-shot translation \citep{johnson2016google}, we hypothesize that performance on the source-original translation task can be improved by distinguishing target-original and target-translationese examples in the training data and constructing an NMT model to perform zero-shot original$\to$original translation.

Because most MT training sets do not annotate each sentence pair's original language, we train a binary classifier to predict whether the target side of a pair is original text in that language or translated from the source language. This follows several prior works attempting to identify translations \citep{Kurokawa09,Koppel11,Lembersky12adapting}.

\begin{figure}
    \centering
    \includegraphics[width=\linewidth]{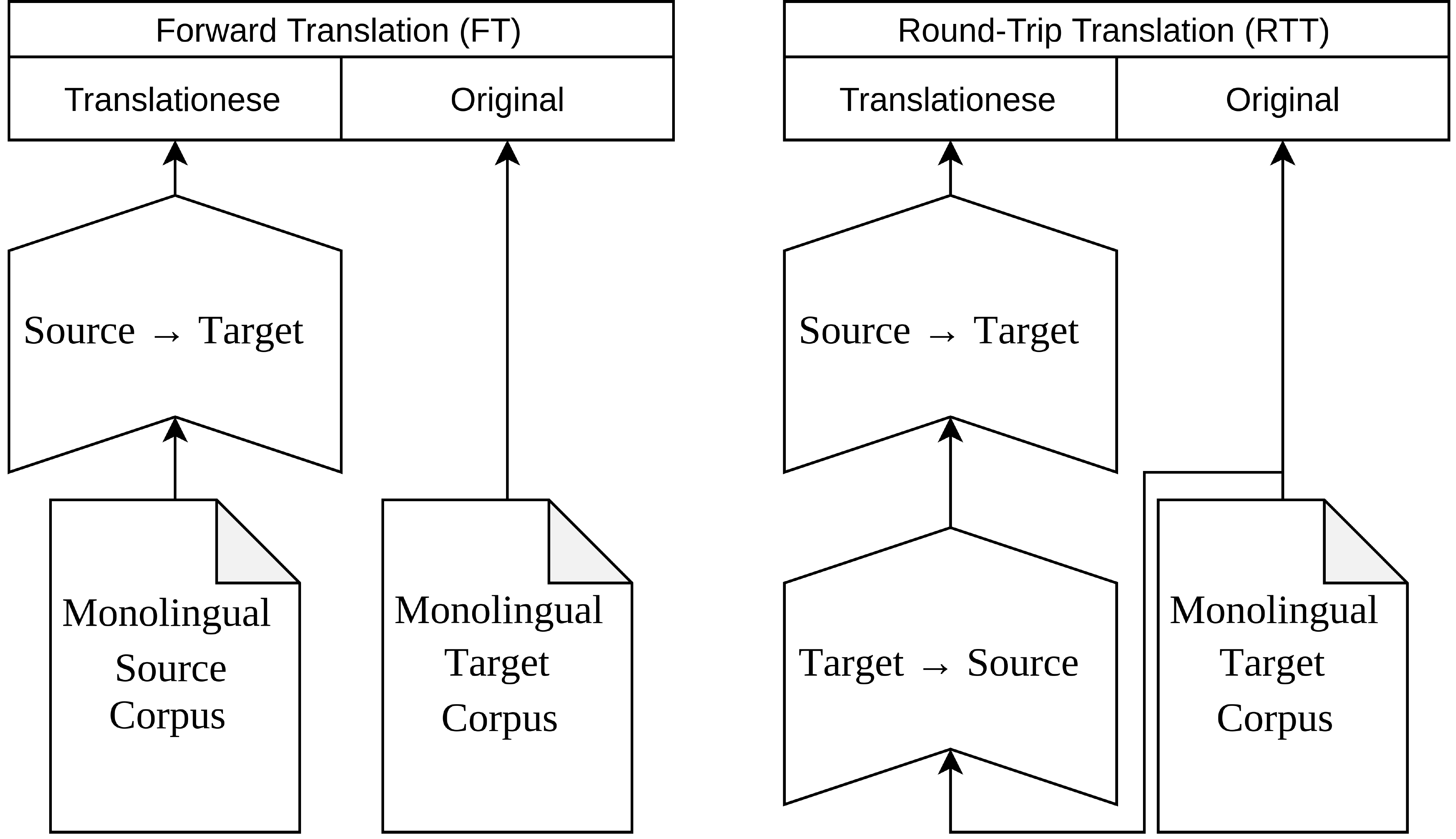}
    \caption{Illustration of data set creation for the FT and RTT translationese classifiers. The Source$\to$Target and Target$\to$Source nodes represent NMT systems.}
    \label{fig:ft_rtt}
\end{figure}

To train the classifier, we need target-language text annotated by whether it is original or translated.
We use News Crawl data from WMT\footnote{http://www.statmt.org/wmt18/translation-task.html} as target-original data. It consists of news articles crawled from the internet, so we assume that most of them are not translations. Getting translated data is trickier; most human-translated pairs where the original language is annotated are only present in test sets, which are generally small. To sidestep this, we choose to use machine translation as a proxy for human translationese, based on the assumption that they are similar. This allows us to create classifier training data using only unannotated monolingual data. We propose two ways of doing this: using forward translation (FT) or round-trip translation (RTT). Both are illustrated in Figure~\ref{fig:ft_rtt}.

To generate FT data, we take source-language News Crawl data and translate it into the target language using a machine translation model trained on WMT training bitext. We can then train a classifier to distinguish the generated text from monolingual target-language text.

One potential problem with the FT data set is that the original and translated pairs may differ not only in the respects we care about (i.e.\ translationese), but also in content. Taking English$\to$French as an example language pair, one could imagine that certain topics are more commonly reported on in original English language news than in French, and vice versa, e.g. news about American or French politics, respectively. The words and phrases representing those topics could then act as signals to the classifier to distinguish the original language.

To address this, we also experiment with RTT data. For this approach we take target-language monolingual data and round-trip translate it with two machine translation models (target$\to$source and then source$\to$target), resulting in another target-language sentence that should contain the same content as the original sentence, alleviating the concern with FT data. Here we hope that the noise introduced by round-trip translation will be similar enough to human translationese to be useful for our downstream task. 

In both settings, we use the trained binary classifier to detect and tag training bitext pairs where the classifier predicted that the target side is original.

\section{Experimental Set-up}
\subsection{Data}
We perform our experiments on WMT18 English$\to$German bitext and WMT15 English$\to$French bitext. We use WMT News Crawl for monolingual data (2007-2017 for German and 2007-2014 for French). We filter out sentences longer than 250 subwords (see Section~\ref{subsection:architecture} for the vocabulary used) and remove pairs whose length ratio is greater than 2. This results in about 5M pairs for English$\to$German. We do not filter the English$\to$French bitext, resulting in 41M sentence pairs.

For monolingual data, we deduplicate and filter sentences with more than 70 tokens or 500 characters. For the experiments described later in Section~\ref{subsection:bt_experiments}, this monolingual data is back-translated with a target-to-source translation model; after doing so, we remove any sentence pairs where the back-translated source is longer than 75 tokens or 550 characters. This results in 216.5M sentences for English$\to$German (of which we only use 24M at a time) and 39M for English$\to$French. As a final step, we use an in-house language identification tool based on the publicly-available Compact Language Detector 2\footnote{https://github.com/CLD2Owners/cld2} to remove all pairs with the incorrect source or target language. This was motivated by observing that some training pairs had the incorrect language on one side, including cases where both sides were the same; \citet{Khayrallah18} found that this type of noise is especially harmful to neural models.

The classifiers were trained on the target language monolingual data in addition to either an equal amount of source language monolingual data machine-translated into the target language (for the FT classifiers) or the same target sentences round-trip translated through the source language with MT (for the RTT classifiers). In both cases, the MT models were trained only with WMT bitext.

The models used to generate the synthetic data have BLEU~\cite{papineni2002bleu} performance as follows on newstest2014/full: German$\to$English 31.8; English$\to$German 28.5; French$\to$English 39.2; English$\to$French 40.6. Here and elsewhere, we report BLEU scores with SacreBLEU~\cite{post2018call}; see Section~\ref{subsection:evaluation}.

Both language pairs considered in this work are high-resource. While translationese is a potential concern for all language pairs, in low-resource settings it is overshadowed by general quality concerns stemming from the lack of training data. We leave for future work the application of these techniques to low-resource language pairs.

\subsection{Architecture and Training}\label{subsection:architecture}
Our NMT models use the transformer-big architecture~\cite{vaswani2017attention} implemented in {\it lingvo}~\cite{shen2019lingvo} with a shared source-target byte-pair-encoding (BPE) vocabulary \citep{sennrich2016neural} of 32k types. To stabilize training, we use exponentially weighted moving average (EMA) decay \cite{buduma2017fundamentals}. Checkpoints were picked by best dev BLEU on a set consisting of a tagged and untagged version of every input.

For the translationese classifier, we trained a three-layer CNN-based classifier optimized with Adagrad. We picked checkpoints by F1 on the development set, which was newstest2015 for English$\to$German and a subset of newstest2013 containing 500 English-original and 500 French-original sentence pairs for English$\to$French. We found that the choice of architecture (RNN/CNN) and hyperparameters did not make a substantial difference in classifier accuracy.

\subsection{Evaluation}\label{subsection:evaluation}
We report BLEU~\cite{papineni2002bleu} scores with SacreBLEU~\cite{post2018call} and include the identification string\footnote{BLEU + case.mixed + lang.LANGUAGE\_PAIR + numrefs.1 + smooth.exp + test.SET + tok.intl + version.1.2.15} to facilitate comparison with future work. We also run human evaluations for the best performing systems (Section~\ref{subsection:human_eval}).

\section{Results and Discussion}\label{section:results}

\subsection{Classifier Accuracy}

\begin{table}[t]
\begin{center}
{\setlength{\tabcolsep}{.51em}
{
\begin{tabular}{ |c||c|c|c|}
  \hline
  Language & Classifier & Bitext & BT   \\
   & Type & \% Orig. & \% Orig.   \\
  \hline \hline
  \multirow{2}{*}{French} & FT & 47\% & 84\% \\
  \cdashline{2-4}
   &  RTT & 30\% & 68\% \\
  \hline
 \multirow{2}{*}{German} & FT & 22\%* & 82\% \\
   \cdashline{2-4}
   & RTT & 29\%* & 70\% \\
  \hline
  
\end{tabular}}}
\end{center}
\caption{Percentage of training data where the target side was classified as original. English$\to$German pairs with predicted original German (marked with a *) were upsampled to balance both bitext subsets' sizes.}
\label{table:classifier_proportions}
\vspace{-1mm}
\end{table}

\begin{table}[t]
\begin{center}
{
{\setlength{\tabcolsep}{.48em}
\begin{tabular}{ |c||c|c|c|c|c|}
  \hline
  Test set $\rightarrow$ & \multicolumn{2}{c|}{Src-Orig} & \multicolumn{2}{c|}{Trg-Orig} & Both   \\
  \hdashline
  Decode $\rightarrow$ & Nt. & Tr. & Tr. & Nt. &  Match    \\ 
  \hdashline
  Match? $\rightarrow$ & \xmark & \cmark & \xmark & \cmark & \cmark   \\
  \hline
\multicolumn{6}{l}{a. En$\to$Fr: Avg.  newstest20\{14/full,15\}}\\
\hline
 	Untagged &  \textbf{39.5} & 39.5 & \textbf{44.5} & 44.5 & 42.0 \\
 \hline
    FT clf. & 37.7 & \textbf{40.0} & 42.5 & \textbf{45.0} & \textbf{42.5} \\
 \hline
    RTT clf.  & 38.0 & 39.4 & 43.2 & 44.1 & 41.8 \\
 \hline
\multicolumn{6}{l}{b. En$\to$De: Avg. newstest20\{14/full,16,17,18\}}\\
 \hline
 	Untagged & \textbf{36.3} & \textbf{36.3} & \textbf{30.0} & 30.0 & \textbf{34.0} \\
 \hline
    FT clf.  & 28.3 & 36.0 & 29.4 & 29.8 & 33.6 \\
 \hline
    RTT clf. & 32.3  & 36.2 & \textbf{30.0} & \textbf{30.2} & 33.9 \\
\hline
\end{tabular}}
}
\end{center}
\caption{Average BLEU for models trained on (a) WMT 2014 English$\to$French bitext and (b) WMT 2018 English$\to$German bitext, tagged according to target side classifier predictions. The tag controls the output domain: translationese (``Tr") or original/natural text (``Nt."). Matching output and test domains (``Match?" row) for both halves (``Both" column) achieves the highest combined BLEU.}
\label{table:bitext-results}
\vspace{-1mm}
\end{table}

Before evaluating the usefulness of our translationese classifiers for the downstream task of machine translation, we can first evaluate how accurate they are at distinguishing original text from human translations. We use WMT test sets for this evaluation, because they consist of source-original and target-original sentence pairs in equal number.

For French, the FT classifier scored $0.81$ F1 and the RTT classifier scored $0.68$ on newstest2014/full. For German, the FT classifier achieved $0.85$ F1 and the RTT classifier scored $0.65$ on newstest2015.  
We note that while the FT classifiers perform reasonably well, the RTT classifiers are less effective. This result is in line with prior work by \citet{Kurokawa09}, who trained an SVM classifier on French sentences to detect translations from English. They used word n-gram features for their classifier and achieved 0.77 F1, but were worried about a potential content effect and so also trained a classifier where nouns and verbs were replaced with corresponding part-of-speech (POS) tags, achieving 0.69 F1. Note that they tested on the Canadian Hansard corpus (containing Canadian parliamentary transcripts in English and French) while we tested on WMT test sets, so the numbers are not directly comparable, but it is interesting to see the similar trends in comparing content-aware and content-unaware versions of the same method. We also point out that \citet{Kurokawa09} both trained and tested with human-translated sentences, while we trained our classifiers with machine-translated sentences while still testing on human-translated data.

The portion of our data classified as target-original by each classifier is reported in Table~\ref{table:classifier_proportions}.

    \begin{table*}[t]
        \centering
        \begin{tabular}{ |c|c|c|c|} 
            \hline
           Test set $\rightarrow$ & \multicolumn{3}{c|}{Src-Orig} \\
           \cline{2-4}
             Tagging $\downarrow$  & Decode & BLEU & \% Preferred \\ \hline \hline
            Untagged &  - & \textbf{43.9} & 26.6\% \\ \hline
            FT clf. & Natural & 41.5 & \textbf{31.9\% }\\ \hline
        \end{tabular}
        \begin{tabular}{ |c|c|c |c |} 
            \hline
             Test set $\rightarrow$ & \multicolumn{3}{c|}{Src-Orig} \\
           \cline{2-4}
             Tagging $\downarrow$  & Decode & BLEU & \% Preferred \\ \hline \hline
            FT clf.  & Transl. & \textbf{44.6 }& 24.2\%  \\ \hline
            FT clf.  & Natural & 41.5 & \textbf{30.7\%} \\ \hline
        \end{tabular}
        \caption{Fluency side-by-side human evaluation for WMT English$\to$French newstest2014/full (Table~\hyperref[table:bitext-results]{2a}). We evaluate only the source-original half of the test set because it corresponds to our goal of original$\to$original translation. Despite a BLEU drop, humans rate the natural decode on average as more fluent than both the bitext model output and the same model with the translationese decode.}
        \label{tab:humaneval_fluency}
        \vspace{-3mm}
    \end{table*}

\subsection{NMT with Translationese-Classified Bitext}

Table~\hyperref[table:bitext-results]{2a} shows the BLEU scores of three models all trained on WMT 2014 English$\to$French  bitext. They differ in how the data was partitioned: either it wasn't, or tags were applied to those sentence pairs with a target side that a classifier predicted to be original French.  We first note that the model trained on data tagged by the round-trip translation (RTT) classifier performs slightly worse than the baseline. However, the model trained with data tagged by the forward translation (FT) classifier is able to achieve an improvement of 0.5 BLEU on both halves of the test set when biased toward translationese on the source-original half and original text on the target-original half. This, coupled with the observation that the BLEU score on the source-original half sharply drops when adding the tag, indicates that the two halves of the test set represent quite different tasks, and that the model has learned to associate the tag with some aspects specific to generating original text as opposed to translationese.

However, we were not able to replicate this positive result on the English$\to$German language pair (Table~\hyperref[table:bitext-results]{2b}). Interestingly, in this scenario the relative ordering of the FT and RTT models is reversed, with the German RTT-trained model outperforming the FT-trained one. This is also interesting because the German FT classifier achieved a higher F1 score than the French one, indicating that a classifier's performance alone is not a sufficient indicator of its effect on translation performance. One possible explanation for the negative result is that the English$\to$German bitext only contains 5M pairs, as opposed to the 41M for English$\to$French, so splitting the data into two portions could make it difficult to learn both portions' output distributions properly.

\subsection{Human Evaluation Experiments}\label{subsection:human_eval}
In the previous subsection, we saw that BLEU for the source-original half of the test set went down 
when the model trained with FT classifications (\textit{FT clf.})\ was decoded it as if it were target-original (Table~\hyperref[table:bitext-results]{2a}). 
Prior work has shown that BLEU has a low correlation with human judgments when the reference contains translationese but the system output is biased toward original/natural text \citep{Freitag19}. This is the very situation we find ourselves in now.
Consequently, we run a human evaluation to see if the output truly is more natural and thereby preferred by human raters, despite the loss in BLEU.
We run both a fluency and an adequacy evaluation for English$\to$French to compare the quality of this system when decoding as if source-original vs.\ target-original. We also compare the system with the \emph{Untagged} baseline.  All evaluations are conducted with bilingual speakers whose native language is French, and each is rated by 3 different raters, with the average taken as the final score. Our two evaluations are as follows:

\begin{itemize}
    \item \textbf{Adequacy}:
        Raters were shown only the source sentence and the model output. Each output was scored on a 6-point scale.
        
    \item \textbf{Fluency}:
        Raters saw two target sentences (two models' outputs) without the source sentence, and were asked to select which was more fluent, or whether they were equally good.
 \end{itemize}
 
Fluency human evaluation results are shown in Table~\ref{tab:humaneval_fluency}. We measured inter-rater agreement using Fleiss' Kappa \citep{Fleiss71}, which attains a maximum value of 1 when raters always agree. This value was 0.24 for the comparison with the untagged baseline, and 0.16 for the comparison with the translationese decodes. The agreement levels are fairly low, indicating a large amount of subjectivity for this task. However, raters on average still indicated a preference for the \textit{FT clf.} model's natural decodes. This provides evidence that they are more fluent than both the translationese decodes from the same model and the baseline untagged model, despite the drop in BLEU compared to each.

Adequacy human ratings are summarised in Table~\ref{tab:humaneval_adequacy}. Both decodes from the \textit{FT clf.} model scored significantly better than the baseline. This is especially true of the natural decodes, demonstrating that the model does not suffer a loss in adequacy by generating more fluent output, and actually sees a significant gain. We hypothesize that splitting the data as we did here allowed the model to learn a sharper distribution for both portions, thereby increasing the quality of both decode types. Some additional evidence for this is the fact that the \textit{FT clf.} model's training loss was consistently lower than that of the baseline.

\begin{center}
    \begin{table}
        \centering
        \begin{tabular}{ |c| c | c | c|} 
            \hline
              Test set $\rightarrow$ & \multicolumn{3}{c|}{Src-Orig} \\
   \cline{2-4}
            Tagging $\downarrow$  & Decode & BLEU &  Adequacy  \\ \hline \hline
            Untagged & - & 43.9 & 4.51 \\ \hline
            FT clf. & Transl. & \textbf{44.6 }& \enspace 4.67* \\ \hline
            FT clf. & Natural & 41.5  & \enspace \enspace \textbf{4.72}** \\ \hline
        \end{tabular}
        \caption{Human evaluation of adequacy for WMT English$\to$French on the source-original half of newstest2014/full. Humans rated each output separately on a 6-point scale. As with fluency (Table \ref{tab:humaneval_fluency}), the natural decode scores the best, despite a BLEU loss. The single and double asterisks indicate that the adequacy value is significantly greater than the first row's value at significance level $\alpha = 0.05$ and $\alpha = 0.01$, respectively, according to a one-tailed paired $t$-test. The difference between the second and third rows was not significant at $\alpha = 0.1$.}
        \label{tab:humaneval_adequacy}
    \vspace{-3mm}
    \end{table}
\end{center}

\section{Supplemental Experiments}\label{section:ablation}

\subsection{Measuring Translationese}

Translationese tends to be simpler, more standardised and more explicit~\cite{Baker93} compared to original text and can retain typical characteristics of the source language~\cite{toury2012descriptive}.
\newcite{Toral19} proposed metrics attempting to quantify the degree of translationese present in a translation.
Following their work, we quantify lexical simplicity with two metrics: lexical variety and lexical density. We also calculate the length variety between the source sentence and the generated translations to measure interference from the source.

\subsubsection{Lexical Variety}
An output is simpler when it uses a lower number of unique tokens/words. By generating output closer to original target text, our hope is to increase lexical variety.
Lexical variety is calculated as the type-token ratio (TTR):
\begin{equation}\label{eq_ttr}
TTR = \frac{number~of~types}{number~of~tokens}
\end{equation}
\subsubsection{Lexical Density} \label{density}
\newcite{scarpa2006corpus} found that translationese tends to be lexically simpler and have a lower percentage of content words (adverbs, adjectives, nouns and verbs) than original written text. Lexical density is calculated as follows:
\begin{equation}\label{eq_lex_density}
lex\_density = \frac{number~of~content~words}{number~of~total~words}
\end{equation}
\subsubsection{Length Variety} \label{section:length_ratio}
Both MT and humans tend to avoid restructuring the source sentence and stick to sentence structures popular in the source language. This results in a translation with similar length to that of the source sentence. By measuring the length variety, we measure interference in the translation because its length is guided by the source sentence's structure. We compute the normalized absolute length difference at the sentence level and average the scores over the test set of source-target pairs $(x,y)$:

\begin{equation}\label{eq_len}
length~variety = \frac{\left| |x| - |y| \right|}{|x|}
\end{equation}
\subsubsection{Results}

Results for all three different translationese measurements are shown in Table~\ref{tab:translationese_measuring}.

\begin{center}
    \begin{table}[h]
        \centering
        {\setlength{\tabcolsep}{.48em}
        \begin{tabular}{ |c|c|c|c|c |c|} 
        \hline
              Test set $\rightarrow$ & \multicolumn{4}{c|}{Src-Orig} \\
   \cline{2-5}
            Tagging $\downarrow$  & Decode &  Lex. & Lex. & Len.\\ 
            & & Var. & Density & Var.\\ \hline \hline
            Untagged & - & 0.258 & 0.393 & 0.246 \\ \hline
            FT clf. & Transl. &  0.255 & 0.396 & \textbf{0.264}\\ \hline
            FT clf. & Natural &  \textbf{0.260} & \textbf{0.397} & 0.245 \\ \hline
        \end{tabular}
        }
        \caption{Measuring the degree of translationese for WMT English$\to$French newstest2014/full on the source-original half. Higher lexical variety, lexical density, and length variety indicate less translationese output.}
        \label{tab:translationese_measuring}
    \end{table}
    \vspace{-3mm}
\end{center}

\textbf{Lexical Variety} :
     Using the tag to decode as natural text (i.e.\ more like original target text) increases lexical variety. This is expected as original sentences tend to use a larger vocabulary.
     
 \textbf{Lexical Density} :
     We also increase lexical density when decoding as natural text. In other words, the model has a higher percentage of content words in its output, which is an indication that it is more like original target-language text.
     
\textbf{Length Variety} :
    Unlike the previous two metrics, decoding as natural text does not lead to a more ``natural'' (i.e. larger) average length variety. One reason may be related to the fact that this is the only metric that also depends on the source sentence: since all of our training pairs feature translationese on either the source or target side, both the tagged and untagged training pairs will feature similar sentence structures, so the model never fully learns to produce different structures. This further illustrates the problem of the lack of original$\to$original training data noted in the introduction. 

\subsection{Tagging using Translationese Heuristics}
Rather than tagging training data with a trained classifier, as explored in the previous sections, it might be possible to tag using much simpler heuristics, and achieve a similar effect. We explore two options here.

\begin{table*}[t]
\begin{center}
{\small
\begin{tabular}{ |c||c|c|c|c|}
  \hline
  Test set $\rightarrow$ & Src-Orig & Src-Orig & Trg-Orig & Trg-Orig   \\
  \hdashline
  Decode as if $\rightarrow$ & Natural & Transl. & Transl. & Natural    \\ 
  \hdashline
  $\therefore$ Domain match? $\rightarrow$ & \xmark & \cmark & \xmark & \cmark   \\
  Train data tagging $\downarrow$ &  &  &  &    \\
  \hline \hline
 	Untagged & \textbf{39.5} & 39.5  & \textbf{44.5} & 44.5   \\
 \hline
    FT clf. & 37.7  & \textbf{40.0} &  42.5 & \textbf{45.0} \\
 \hline
    Length Variety & 38.2 & 36.1  & 43.6 & 36.2\\
 \hline
    Lex. Density & 36.9 & 36.7& 41.2  & 43.4 \\
 \hline
  
\end{tabular}}
\end{center}
\caption{Comparing heuristic- and classifier-based tagging. BLEU scores are averaged for newstest2014/full and newstest2015 English$\to$French. 
The trained classifier outperforms both heuristics, and length-ratio tagging has the reverse effect from what we expect.}
\vspace{-3mm}
\label{table:heuristics}
\end{table*}

\subsubsection{Length Ratio Tagging}
Here, we partition the training pairs $(x, y)$ according to a simple length ratio $\frac{|x|}{|y|}$. We use a threshold $\hat \rho_{length}$ empirically calculated from two large monolingual corpora, $M_x$ and $M_y$:
\begin{equation}
    \hat \rho_{length} = \frac{\frac{1}{|M_x|}\sum_{x_i \in M_x}|x_i|}{ \frac{1}{|M_y|} \sum_{y_i \in M_y}|y_i|}
\end{equation}
    For English$\to$French, we found $ \hat \rho_{length} = 0.8643$, meaning that original French sentences tend to have more tokens than English. We tag all pairs with length ratio greater than $\hat \rho_{length}$ (49.8\% of the training bitext). Based on the discussion in Section~\ref{section:length_ratio}, we expect that $\frac{|x|}{|y|} \approx 1.0$ indicates translationese, so in this case the tag should mean ``produce translationese" instead of ``produce original text."
    
\subsubsection{Lexical Density Tagging}
We tag examples with a target-side lexical density of greater than 0.5, which means that the target is more likely to be original than translationese. Please refer to Section \ref{density} for an explanation of this metric.

\subsubsection{Results}
Table \ref{table:heuristics} shows the results for this experiment, compared to the untagged baseline and the classifier-tagged model from Table~\hyperref[table:bitext-results]{2a}. This table specifically looks at the effect of controlling whether the output should feature more or less translationese on each subset of the test set.
We see that the lexical density tagging approach yields expected results, in that the tag can be used to effectively increase BLEU on the target-original portion of the test set. The length-ratio tagging, however, has the opposite effect: producing shorter outputs (``decode as if translationese'') produces higher target-original BLEU and lower source-original BLEU. We speculate that this data partition has accidentally picked up on some artifact of the data.

Two interesting observations from Table \ref{table:heuristics} are that 1) both heuristic tagging methods perform much more poorly than the classifier tagging method on both test set halves, and 2) all varieties of tagging produce large performance changes (up to -7.2 BLEU). This second observation highlights that tagging can be powerful -- and dangerous when it does not correspond well with the desired feature.

\subsection{Back-Translation Experiments}\label{subsection:bt_experiments}
We also investigated whether using a classifier to tag training data improved model performance in the presence of back-translated (BT) data. \citet{Caswell19} introduced tagged back-translation (TBT), where all back-translated pairs are tagged and no bitext pairs are. They experimented with decoding the model with a tag (``as-if-back-translated'') but found it harmed BLEU score. However, in our early experiments we discovered that doing this actually \textit{improved} the model's performance on the target-original portion of the test set, while harming it on the source-original half.
Thus, we frame TBT as a heuristic method for identifying target-original pairs: the monolingual data used for the back-translations is assumed to be original, and the target side of the bitext is assumed to be translated. We wish to know whether we can find a better tagging scheme for the combined BT+bitext data, based on a classifier or some other heuristic.

\begin{table*}[t]
\begin{center}
{\small
\begin{tabular}{ |c|c||c|c|c|c|c|}
  \hline
   \multicolumn{2}{|c||}{Test set $\rightarrow$ } & \multicolumn{2}{c|}{Src-Orig} & \multicolumn{2}{c|}{Trg-Orig} & Combined   \\
  \hdashline
  \multicolumn{2}{|c||}{Decode as if $\rightarrow$ }& Natural & Transl. & Transl. & Natural &  Both    \\ 
  \hdashline
  \multicolumn{2}{|c||}{$\therefore$ Domain match? $\rightarrow$ } & \xmark & \cmark & \xmark & \cmark & \cmark   \\
  Bitext tagging $\downarrow$ & BT tagging $\downarrow$ &  &  &  &  &  \\
  \hline
  \multicolumn{7}{l}{}\\
  \multicolumn{7}{l}{a. English$\to$French: Avg. newstest20\{14/full, 15\}}\\
  \hline
    Untagged & All Tagged & 38.4 & 40.8 & 47.5 & 49.8 & 45.5 \\
 \hline
    FT clf. & All Tagged & \textbf{38.8} & 40.8 & 47.3 & \textbf{50.3} & \textbf{45.7} \\
 \hline
    FT clf. & FT clf. & 38.2 & \textbf{40.9} & 45.5 & 49.0 & 45.2 \\
 \hline
    RTT clf. &  RTT clf. & 38.3 & 40.1 & \textbf{49.4} & 49.5 & 45.1  \\
 \hline
  \multicolumn{7}{l}{}\\
 \multicolumn{7}{l}{b. English$\to$German: Avg. newstest20\{14/full,16,17,18\}}\\
 \hline
    Untagged & All Tagged & 33.5 & 37.3 & 36.7 & 37.1 & \textbf{37.6} \\
 \hline
    FT clf. & All Tagged & 33.4 & 37.2 & 36.2 & \textbf{37.2} & 37.5 \\
 \hline
    RTT clf. & All Tagged & \textbf{33.6} & \textbf{37.4} & 36.6 & 37.1 & \textbf{37.6} \\
 \hline
    RTT clf. & RTT clf. & 31.6 & 35.7 & \textbf{36.8} & 36.7 & 36.4 \\
 \hline
    FT clf. & FT clf. & 30.5 & 35.5 & 36.5 & 37.0 & 36.5 \\
 \hline
\end{tabular}}
\end{center}
\caption{Average BLEU scores for models trained on (a) WMT 2018 English$\to$French bitext plus 39M back-translated monolingual sentences, and (b) WMT 2018 English$\to$German bitext plus 24M back-translated monolingual sentences. As before, we tag by heuristics and/or classifier predictions on the target (German) side.
}
\label{table:bt-results}
\end{table*}

Results for English$\to$French models trained with BT data are presented in Table~\hyperref[table:bt-results]{7a}. While combining the bitext classified by the FT classifier with all-tagged BT data yields a minor gain of 0.2 BLEU over the TBT baseline of \citet{Caswell19}, the other methods do not beat the baseline. This indicates that assuming all of the target monolingual data to be original is not as harmful as the error introduced by the classifiers.

English$\to$German results are presented in Table~\hyperref[table:bt-results]{7b}. Combining the bitext classified by the RTT classifier with all-tagged BT data matched the performance of the TBT baseline, but none of the models outperformed it. This is expected, given the poor performance of the bitext-only models for this language pair.

\section{Example Output}
In Table~\ref{tab:example_output}, we show example outputs for WMT English$\to$French comparing the \emph{Untagged} baseline with the \emph{FT clf.} natural decodes.
In the first example, \textit{avec suffisamment d'art} is an incorrect word-for-word translation, as the French word \textit{art} cannot be used in that context. Here the word \textit{habilement}, which is close to ``skilfully" in English, sounds more natural.
In the second example, \textit{libre d'imp{\^ o}t} is the literal translation of ``tax-free", but French documents rarely use it, they prefer \textit{pas imposable}, meaning ``not taxable".

\begin{table*}[ht]
    \centering
    \setlength\tabcolsep{4pt}
    \begin{tabular}{c|l}
    \hline
    Source & Sorry she didn't phrase it artfully enough for you. \\
    Untagged & D{\'e}sol{\'e}e, elle ne l'a pas formul{\'e} \textbf{avec suffisamment d'art} pour vous. \\ 
    FT clf. & D{\'e}sol{\'e} elle ne l'a pas formul{\'e} assez \textbf{habilement} pour vous. \\ \hline
    Source & Your first £10,000 is \textbf{tax free}. \\
    Untagged & Votre premi{\`e}re tranche de 10 000 £ est \textbf{libre d'imp{\^ o}t}. \\
    FT clf. & La premi\`ere tranche de 10 000 £ n'est \textbf{pas imposable}. \\ \hline
    \end{tabular}
    \caption{Example English$\to$French output comparing the untagged baseline with the \emph{FT clf.} natural decode.}
\label{tab:example_output}
\end{table*}

\section{Related Work}
\subsection{Translationese}
The effects of translationese on MT training and evaluation have been investigated by many prior authors \citep{Kurokawa09,Lembersky12adapting,Toral18,Zhang19,Graham19,Freitag19,Edunov19,freitag2020bleu}. Training classifiers to detect translationese has also been done \citep{Kurokawa09,Koppel11,shen2019sourcetarget}. Similarly to this work, \citet{Kurokawa09} used their classifier to preprocess MT training data; however, they completely removed target-original pairs. In contrast, \citet{Lembersky12adapting} used both types of data (without explicitly distinguishing them with a classifier), and used entropy-based measures to cause their phrase-based system to favor phrase table entries with target phrases that are more similar to a corpus of translationese than original text. In this work, we combine aspects from each of these: we train a classifier to partition the training data, and use both subsets to train a single model with a mechanism allowing control over the degree of translationese to produce in the output. We also show with human evaluations that source-original test sentence pairs result in BLEU scores that do not correlate well with translation quality when evaluating models trained to produce more original output.

\subsection{Training Data Tagging for NMT}
In addition to the methods in \citet{Caswell19}, tagging training data and using the tags to control output is a technique that has been growing in popularity. Tags on the source sentence have been used to indicate target language in multilingual models~\cite{johnson2016google}, formality level in English$\to$Japanese \cite{yamagishi2016controlling}, politeness in English$\to$German \cite{Sennrich16controlling}, gender from a gender-neutral language \cite{kuczmarski2018gender}, as well as to produce domain-targeted translation \citep{kobus2016controlling}. \citet{shu2019generating} use tags at training and inference time to increase the syntactic diversity of their output while maintaining translation quality; similarly, \citet{agarwal2019controlling} and \citet{Marchisio19} use tags to control the reading level (e.g. simplicity/complexity) of the output.
Overall, tagging can be seen as domain adaptation~\cite{freitag2016fast,luong2015stanford}.

\section{Conclusion}
We have demonstrated that translationese and original text can be treated as separate target languages in a ``multilingual'' model, distinguished by a classifier trained using only monolingual and synthetic data. The resulting model has improved performance in the ideal, zero-shot scenario of original$\to$original translation, as measured by human evaluation of adequacy and fluency. However, this is associated with a drop in BLEU score, indicating that better automatic evaluation is needed.

\paragraph{Acknowledgments}
We are grateful to the anonymous reviewers for suggesting useful additions.
\bibliography{all}
\bibliographystyle{acl_natbib}
\end{document}